\documentclass[10pt,twocolumn,letterpaper]{article}

\usepackage[pagenumbers]{cvpr} %

\usepackage{currfile}
\usepackage[normalem]{ulem}  %
\usepackage{enumitem}  %
\usepackage{colortbl}
\usepackage{afterpage}
\usepackage{nicefrac}
\usepackage{siunitx}
\usepackage[hang,flushmargin,symbol]{footmisc}
\usepackage{blindtext}
\usepackage{multirow}
\usepackage{siunitx}
\usepackage{lipsum}
\usepackage[figuresleft]{rotating}
\usepackage{svg}
\usepackage{adjustbox}
\usepackage{soul}
\usepackage{nicefrac}
\usepackage{threeparttable}
\usepackage{makecell}
\usepackage{float}
\usepackage{colortbl}
\usepackage{balance}
\usepackage{csquotes}

\usepackage{pifont}

\def \iwalt {I-WALT\xspace}

\def \vwalt {V-WALT\xspace}

\def \Timagenet {\emph{ImageNet}\xspace}
\def \Tplaces {\emph{Places365}\xspace}
\def \Tinat {\emph{Inat2018}\xspace}
\def \Tkfour {\emph{K400}\xspace}
\def \Tkseven {\emph{K700}\xspace}
\def \Tscannet {\emph{Depth}\xspace}
\def \Twaymo {\emph{Obj. Tracks}\xspace}
\def \Tssv {\emph{SSv2}\xspace}
\def \Tpt {\emph{PointTracks}\xspace}
\def \Tpose {\emph{Cam. Pose}\xspace}

\let\oldparagraph\paragraph
\renewcommand{\paragraph}[1]{\oldparagraph{#1.}}

\definecolor{cvprblue}{rgb}{0.21,0.49,0.74}
\usepackage[pagebackref, breaklinks,colorlinks, allcolors=cvprblue]{hyperref}

\usepackage[capitalize]{cleveref}
\crefname{section}{Sec.}{Secs.}
\Crefname{section}{Section}{Sections}
\Crefname{table}{Table}{Tables}
\crefname{table}{Tab.}{Tabs.}

\def\paperID{3205}
\def\confName{CVPR}
\def\confYear{2025}

\title{
From Image to Video:
An Empirical Study of Diffusion Representations
}

\author{
{
\hypersetup{
    colorlinks=true,
    linkcolor=black,
    filecolor=black,
    urlcolor=black,
    citecolor=black,
}
\fontsize{10.5}{12.6}\selectfont 
\href{https://www.linkedin.com/in/pdvelez/en}{Pedro Vélez}\footnotemark[1]\protect\phantom{\footnotesize 1}\footnotemark[2] \;
\href{https://scholar.google.com/citations?user=HGLobX4AAAAJ&hl=en}{Luisa F.\ Polanía}\footnotemark[1]\protect\phantom{\footnotesize 1}\footnotemark[2] \;
\href{https://yangyi02.github.io/}{Yi Yang} \;
\href{https://scholar.google.com/citations?user=ep_nM5sAAAAJ&hl=en}{Chuhan Zhang} \;
\href{https://scholar.google.com/citations?user=NVD-BU4AAAAJ&hl=en}{Rishabh Kabra} \;
\href{https://anuragarnab.github.io/}{Anurag Arnab} \;
\href{https://msajjadi.com/}{Mehdi S.\ M.\ Sajjadi}\footnotemark[2]
}
\\[2mm]
{
\hypersetup{
    colorlinks=true,
    linkcolor=gdm,
    filecolor=gdm,
    urlcolor=gdm,
    citecolor=gdm,
}
\fontsize{10.5}{12.6}\selectfont 
\definecolor{gdm}{RGB}{83,131,236}
\href{https://deepmind.google/}{\vphantom{Google DeepMind}\includegraphics[height=12pt]{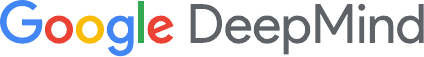}}
}
}
\begin{document}
\maketitle
\begin{abstract}

\footnotetext[0]{
\mbox{
\footnotemark[1]Equal contribution.
\footnotemark[2]Corresponding authors:
\href{mailto:pedrovelez@google.com,polania@google.com,i2v@msajjadi.com}{pedrovelez@google.com},
\href{mailto:pedrovelez@google.com,polania@google.com,i2v@msajjadi.com}{polania@google.com},
\href{mailto:pedrovelez@google.com,polania@google.com,i2v@msajjadi.com}{i2v@msajjadi.com}.
}
}

Diffusion models have revolutionized generative modeling, enabling unprecedented realism in image and video synthesis.
This success has sparked interest in leveraging their representations for visual understanding tasks.
While recent works have explored this potential for image generation, the visual understanding capabilities of video diffusion models remain largely uncharted.
To address this gap, we systematically compare the same model architecture trained for video versus image generation, analyzing the performance of their latent representations on various downstream tasks including image classification, action recognition, depth estimation, and tracking.
Results show that video diffusion models consistently outperform their image counterparts, though we find a striking range in the extent of this superiority.
We further analyze features extracted from different layers and with varying noise levels, as well as the effect of model size and training budget on representation and generation quality.
This work marks the first direct comparison of video and image diffusion objectives for visual understanding, offering insights into the role of temporal information in representation learning.
\end{abstract}

\section{Introduction}
\label{sec:intro}

The field of computer vision has made remarkable strides towards imbuing machines with the ability to see and interpret the visual world.
A key challenge in realizing this goal lies in learning effective representations that capture both the rich semantic information and dynamic 4D structure (3D \& motion) inherent in real-world visual data.
Leading approaches in visual representation learning have primarily focused on \emph{contrastive learning} and \emph{reconstruction}.
Contrastive models can be exemplified by DINO \cite{caron2021emerging} which applies knowledge distillation between augmented views of images, and CLIP \cite{radford2021} which aligns image and text representations, while MAE \cite{mae} and I-JEPA \cite{ijepa} are commonly used reconstruction models that predict masked pixels and features, respectively.

\begin{figure}[t]
\centering
\includegraphics[
trim={2mm 5mm 2mm -12mm},clip,
width=\linewidth,
]{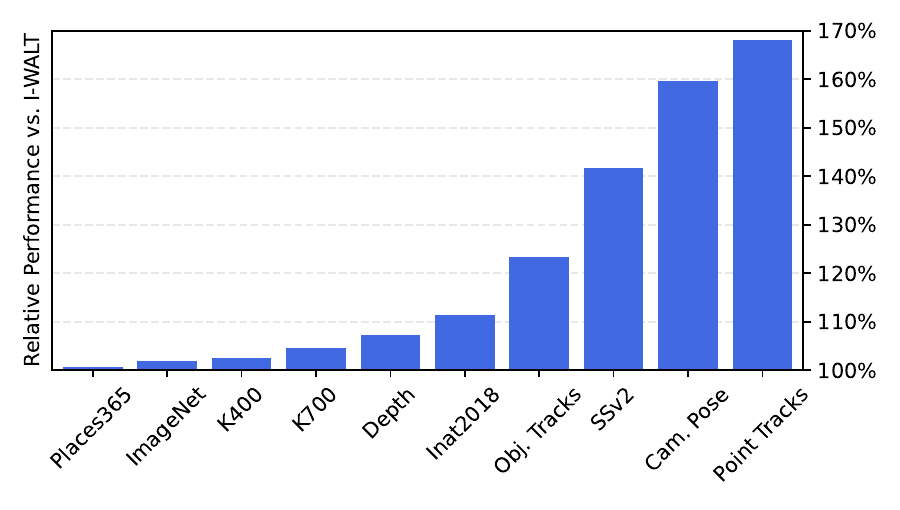}
\setlength{\belowcaptionskip}{-5pt}
\caption{
    \textbf{Video \vs image diffusion representations} --
    The diffusion model \vwalt trained on generating videos learns better features than the same model trained on generating images (\iwalt, normalized to 100\% here) as measured across a range of readout tasks.
    See \cref{sec:exp:imagevsvideo} for details.
}
\label{fig:teaser}
\end{figure}

Meanwhile, generative models have demonstrated an unprecedented ability to synthesize novel, photo-realistic imagery \cite{karras2022elucidating, liu2024sora}.
Among generative approaches, diffusion models \cite{HoDenoisingDiffusionProbabilisticModels} have emerged as the state-of-the-art for both image and video generation, achieving remarkable results in synthesizing high-quality visual content \cite{fuest2024diffusion}.
Image diffusion models have also demonstrated their capacity to learn powerful representations for downstream tasks such as image classification \cite{li2023your}, depth estimation \cite{shao2023monodiffusion} and semantic keypoint matching \cite{hedlin2024unsupervised, tang2023}, suggesting that the denoising process enables these models to acquire a deep understanding of visual semantics and structure.
However, despite the recent emergence of video diffusion models capable of generating high-fidelity video content, the representational power of these models remains largely unexplored, particularly in the context of dynamic scene understanding.
This leaves a crucial question unanswered: how effectively do video diffusion models capture the interplay between motion and spatial scene understanding, and how do they perform compared to image diffusion models?

\clearpage
In this work, we delve into the representational power of the diffusion model WALT \cite{walt} which is particularly well-suited for a direct comparison thanks to its hybrid architecture, allowing us to specifically analyze the implications of the video versus image pre-training objective on downstream performance.

Our key contributions are as follows:
\begin{enumerate}
    \item We compare the same model architecture trained on image \vs video generation and demonstrate the superiority of video for learning representations across diverse downstream tasks.
    \item We provide a qualitative analysis of video diffusion representations, highlighting the role of motion in the video representation space.
    \item We present findings on the relationship between training budget, model size, visual generation quality and downstream performance of the learned representations.
\end{enumerate}
To the best of our knowledge, this is the first work to analyze and compare the internal representations of the same diffusion model pre-trained for video versus image generation across a diverse range of tasks.

\section{Related Work}
\label{sec:related}

Diffusion models \cite{HoDenoisingDiffusionProbabilisticModels} have revolutionized the fields of image and video synthesis as they offer unprecedented generation quality \cite{karras2022elucidating, liu2024sora}.
Whereas they once relied heavily on U-Nets \cite{ronneberger2015u}, transformer-based diffusion models have since gained traction \cite{dit}.
We refer the reader to \citet{zhang2023survey} for a survey on diffusion-based image generation.

\vspace{4mm}
\noindent \textbf{Image diffusion for visual understanding.} Beyond their impressive generative capabilities, image diffusion models have proved remarkably effective in perception tasks.
For example, DDPM-Seg \cite{baranchuk2022} utilizes intermediate features of a U-Net-based diffusion model and excels at semantic segmentation.
It has further been extended to different architectures and showed that the best-performing features are found in the middle upsampling layers using small noise levels \cite{xiang2023}.
Pre-trained text-to-image diffusion models have also been applied to panoptic segmentation \cite{xu2023}.

In terms of correspondence tasks, DIFT~\cite{tang2023}, a method based on extracting features from diffusion models, showed that features extracted in earlier layers with larger timesteps tend to be more semantically meaningful, whereas lower-level features with smaller timesteps focus more on low-level details.
A related work found that merging diffusion representations with DINOv2 \cite{oquab2024} features improves performance on semantic correspondence \cite{zhang2024}.
\citet{luo2023dhf} demonstrated that having an aggregation network that learns weights for all features maps across all layers and timesteps performs better than manually selecting layers and timesteps. 
\citet{clark2024text} proposed a mechanism that makes use of the denoising scores for all possible labels,
thereby creating a zero-shot ImageNet classifier which is however computationally impractical.
Another work compares various types of readout architectures with similar parameter count and has found attentive readout to perform best \cite{mukhopadhyay2023}.

Further works propose diffusion models as a general foundational model for a suite of image understanding tasks.
For example, \citet{zhao2023} showed that diffusion models excel at both semantic segmentation and depth estimation, while \citet{yang2023}, showed state-of-the-art performance on several image classification, semantic segmentation and landmark detection benchmarks using diffusion models.
We refer the reader to a recent survey on the connection between diffusion models and image representation learning for an extensive overview \cite{fuest2024diffusion}.

\noindent \textbf{Video representation learning.} Building on image representation learning, research is naturally progressing to video analysis for enhanced spatial understanding \cite{parthasarathy2023self}.
Contrastive and masked modeling are popular choices here:
V-JEPA \cite{vjepa} predicts video embeddings from a set of random crops in a self-supervised student-teacher setup, while VideoMAE \cite{tong2022videomae, feichtenhofer2022masked, videomaev2} pre-trains vision transformers by masking and reconstructing random video patches.
VideoPrism \cite{zhao2024} combines both: it uses video-language contrastive learning in a first stage to capture semantic content, while a second stage uses masked modeling with global-local distillation and token shuffling.

Recent works have started to explore the potential of \emph{image} diffusion models on video understanding \cite{chen2023, nag2023}.
Since high-quality \emph{video} generation with diffusion models is only a recent development \cite{walt, liu2024sora}, the literature on investigating their features is sparse:
video diffusion representations have recently been found to produce temporally consistent video depth estimates \cite{hu2024depthcrafter}, and they have been applied to object segmentation \citet{zhu2024}, though without a direct comparison with image diffusion models.
A concurrent work with ours, \cite{anonymous2024}, evaluates the performance of the video diffusion model SVD \cite{blattmann2023stable} on video understanding tasks, and finds that it outperforms image baselines, in particular Stable Diffusion (SD) \cite{rombach2022high}.
While SVD builds upon SD's U-Net architecture by incorporating temporal convolution and attention layers, it nearly doubles the parameter count (865\,M to 1.5\,B) which weakens the expressiveness of the comparison.
In this work, we specifically choose WALT \cite{walt} for our investigations due to its hybrid architecture that allows a more apt comparison.
Finally, also concurrently to our work, \citet{man2024} evaluate image and video representation models with a focus on 3D scene understanding (semantic and geometric understanding) and vision-language reasoning tasks and find
diffusion models to excel at geometric tasks.

\section{Method}
\label{sec:method}

We lead this section with a recap of latent diffusion models and the WALT model in \cref{sec:method:latentdiffusionmodels} and \cref{sec:method:walt}, respectively.
\Cref{sec:method:walt_variants} describes how WALT operates on images \vs video, followed by a description of the probing framework for quantitative evaluations in \cref{sec:method:probing_framework}.

\subsection{Latent Diffusion Models}
\label{sec:method:latentdiffusionmodels}

Diffusion Models \cite{sohldickstein2015deepunsupervisedlearningusing} are probabilistic generative models that can learn a distribution by denoising a normally distributed variable. They are based on two stages, a forward diffusion stage and a backward diffusion stage. In the forward diffusion stage, the input data is gradually corrupted by adding noise with a fixed variance schedule until the data distribution degrades into a standard Gaussian distribution.  In the backward stage, the model reconstructs the original input data by learning to gradually denoise the signal.

Latent Diffusion Models (LDMs) \cite{blattmann2023stable}  apply the diffusion process in the latent space of a Vector Quantized Variational Autoencoder (VQ-VAE) \cite{NeuralDiscreteRepresentationLearning, TamingTransformersForHiResImageSynth} which helps to significantly reduce the computation requirements when compared to using raw pixels. The VQ-VAE is composed of an encoder $E(x)$ that encodes a video or an image ${x}\,{\in}\,\mathbb{R}^{T{\times}H{\times}W{\times}3}$ into a compressed latent representation $ z \in \mathbb{R}^{t\times h \times w \times c} $ and a decoder $D$ that reconstructs the input data from the latent $\widetilde{x}=D(z)$. 

The inverse diffusion process is modeled by applying a learnable function $f_\theta(z_t, t)$ to the noised latents at each step to recover the original input. More formally, $f_\theta(z_t) \approx \nabla \log p(z_t)$, where $p(z_t)$ is the probability density function of the latent space at step $t$,  $z_t = \sqrt{\alpha(t)}z_0 + \sqrt{1-\alpha(t)}\epsilon$ is a noisy version of $z_0$, $\epsilon  \sim  \mathcal{N}(\mathbf{0}, {\mathbf{I}})$, $t \in [0, 1]$, and $\alpha(t)$ is a predefined monotonically decreasing function from 1 to 0.

The function $f_\theta$ is parameterized by a neural network, which is trained with the denoising objective defined as 
\begin{equation*}
\mathbb{E}_{z{\sim}p_{data}, t\sim \mathcal{U}(0, 1), \epsilon \sim \mathcal{N}(\mathbf{0}, \mathbf{I})} \big[\|\epsilon - f_{\theta} (z_t; C, t)\|^2\big],
\end{equation*}
where $C$ is the condition, \eg, class labels or text prompts.

Beginning with a noise sample $z_1 \sim \mathcal{N}(\mathbf{0}, \mathbf{I})$, an iterative sampling process repeatedly applies the model $f_\theta(z_t; C, t)$ to progressively refine an estimate of a clean latent sample $\hat{z}_0$.  This refined latent sample is then decoded, transforming it back into the pixel space.
In this work, we do not make use of the decoder since we extract features from the main transformer module.

\subsection{Windowed-Attention Latent Transformer}
\label{sec:method:walt}

Windowed-Attention Latent Transformer (WALT) \cite{walt}, a transformer-based video diffusion model conditioned on text prompts, is selected for this study because the same architecture can be used for both image and video generation, leading to a fair comparison. Although, a study with a single diffusion model is not ideal, the same fair comparison is not feasible with open-sourced video diffusion models given the architecture size disparities with their image counterparts. For example, there is a significant architecture size disparity between Stable Image Diffusion 2.1 (SD 2.1)~\cite{rombach2022high} and its temporal extension, Stable Video Diffusion (SVD)~\cite{blattmann2023stable}. More precisely, SD 2.1 uses a U-Net architecture with 865 million parameters, while SVD  inserts temporal convolution and attention layers after every spatial convolution and attention layer of SD 2.1, which leads to an additional 656 million parameters.

\begin{figure}[t]
\centering
\includegraphics[
width=\linewidth]{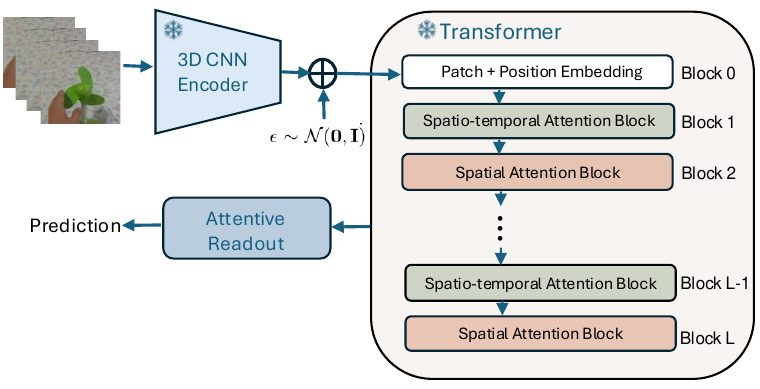}
\setlength{\belowcaptionskip}{-10pt}
\caption{
\textbf{Probing architecture} --
    We feed videos through the model and extract (frozen) intermediate features.
    Cross-attention modules then read out the label for the downstream tasks.
}
\label{fig:architecture}
\end{figure}

WALT is used as a frozen backbone to train light-weight readout heads for downstream perception tasks. It leverages a causal 3D CNN encoder of the MAGVIT-v2 tokenizer \cite{magvitv2} to jointly compress images and videos into a shared latent space, which allow the model to be trained on massive image-text and video-text datasets.

The input to the model is a batch of latent tensors $ z \in \mathbb{R}^{(1+m)\times h \times w \times c} $, generated by the 3D CNN encoder, which are first passed through an embedding block, referred to as Block 0, to be independently encoded as a sequence of non-overlapping patches along with learned position embeddings \cite{AttentionIsAllYouNeed}.
The first frame is encoded independently from the remaining video frames, allowing static images to be treated as videos with a single frame. In particular, the WALT checkpoints used in this paper were shared by the authors and were trained to process sequences of 17 frames.  The video frames are tokenized by the 3D CNN encoder with 5 temporal latents, where the first latent represent the initial frame and the remaining $m=4$ represent the remaining 16 frames. In terms of the spatial compression factor of the latents, it is set to 8 for both width and height.

The WALT architecture introduces a design variation for the transformer in order to reduce the high compute and memory cost of processing image and video tokens. The variation consists of computing self-attention in windows instead of using the traditional global self-attention modules. More precisely, the transformer consists of a concatenation of $L$ window-restricted attention blocks, alternating between spatio-temporal-window blocks and spatial-window blocks. For the case of images, the spatio-temporal-window blocks use an identity attention mask, ensuring that any given latent only attend to itself. This architectural choice enables joint training, where the spatial blocks independently process images and video frames, while the spatio-temporal blocks model motion and temporal dynamics in videos. Also, in terms of inference, it enables both image and video generation modes. 

The WALT model is trained on conditional information such as text embeddings, previously generated samples, and past frames for auto-regressive generation of videos.
While the original WALT model is a cascaded diffusion model with super-resolution stages, we only investigate the base model that generates videos at low resolution in this work.

\subsection{WALT for images and video}
\label{sec:method:walt_variants}
WALT can be used in both video generation, referred to as \vwalt, and image generation modes. When used for image generation, any given latent in the spatio-temporal blocks only attend to itself. For comparison purposes, one drawback of such design is that not only the temporal attention is removed but also the spatial attention. Given that the spatio-temporal blocks in \vwalt perform both spatial and temporal attention, it is fairer to compare with an image counterpart of the WALT model where window-restricted spatio-temporal attention blocks are replaced by window-restricted spatial attention blocks of the same number of parameters. In that way, we can directly measure the impact of adding temporal attention. Such image counterpart was trained for this paper and is referred to as \iwalt. Note that the window-restricted spatial and spatio-temporal blocks only differ in the windows sizes, and therefore, \iwalt and \vwalt share the same architecture.

For training \vwalt, an internal dataset composed of images and videos was used. The same dataset was employed for training \iwalt, but instead of using full videos, frames were randomly extracted from each video. The training settings of \iwalt are the same as WALT.

\subsection{Probing Framework}
\label{sec:method:probing_framework}

A probing framework is introduced to extract video representations from the WALT model and subsequently apply a task-specific readout head for various video understanding tasks. The process starts by adding noise to the latent representations of the input data at a time step $t$ to simulate the forward diffusion process before passing them through the denoiser model. Only a single forward pass of the input through the diffusion model is necessary to extract its visual representation, as opposed to going through the entire multi-step generative diffusion process. The forward pass uses a null text embedding. Subsequently, activations of the transformer intermediate blocks are extracted to train task-specific readout heads that will interpret the learned representations. A diagram illustrating the probing framework is shown in \cref{fig:architecture}.
In order to determine an adequate timestep $t$ and the most representative activation block $l$, we run ablations described in \cref{sec:exp:featureextraction} and showcased in \cref{fig:noise_and_blocks}.

The WALT model is evaluated on representative visual perception tasks, ranging from pure semantics to spatio-temporal understanding: image classification, action recognition, monocular depth estimation, relative camera pose prediction to visual correspondence.
For the evaluation on visual perception tasks, we largely follow the probing methodology of \citet{s4dpaper}.

\paragraph{Image classification}
Image classification, characterized by its purely semantic nature, is one of the most fundamental areas in computer vision. Within this area, the tasks of object classification, scene recognition and fine-grained visual classification are selected for downstream evaluation of WALT. 
An attentive readout is used for this task \cite{vjepa}. The cross-attention layers are trained with a learnable query token, and the output of the cross-attention is added to the query token and then passed to a two-layer MLP with GeLU activation, followed by layer normalization and a linear classification layer.
The readout is trained with the softmax cross-entropy loss.
\Timagenet~\cite{imagenet} and \Tplaces~\cite{places365} are used for object and scene classification, respectively. For fine-grained visual classification we use iNaturalist 2018 (\Tinat)~\cite{inaturalist} which contains visually similar plant and animal species.
Top-1 classification accuracy is used for all the image classification tasks.

\paragraph{Action recognition}
Understanding actions in videos often requires capturing temporal dependencies between frames, \ie the model needs to understand how actions unfold over time and how earlier frames relate to later ones to accurately classify the action.
As above, we use attentive readout, though over all video frames here.
Kinetics-400 and 700 (\Tkfour, \Tkseven)~\cite{kinetics400, k700} are used for appearance-focused action recognition.
For more motion-sensitive action recognition, Something-Something v2 (\Tssv)~\cite{ssv2} is used.
We report top-1 classification accuracy.

\paragraph{Monocular depth prediction}
Monocular depth estimation, referred hereafter as \Tscannet, is a 3D perception task that aims at predicting the distance of surface elements in the scene from the camera. Unlike traditional geometric correspondence and triangulation techniques, this requires only a single image. However, it can also be calculated from video to leverage temporal dependencies between frames. Monocular depth estimation is a fundamental problem in computer vision as it bridges the gap between 2D images and the 3D world.
For the readout, we use the decoder of the Scene Representation Transformer~\cite{SRT} which is composed of a small number
of cross-attention layers followed by an MLP.
Fourier positional encoding~\cite{nerf} applied to the input latents is used to generate a set of queries for the decoder. Each depth pixel value is decoded independently by a transformer that crossattends from a query into the latent features generated by the pretrained model, thereby aggregating relevant information from the latents to predict depth. For training the readout, an L2 loss between the prediction and the ground truth depth map is used. 
We use ScanNet~\cite{scannet}, a dataset of RGB-D videos of indoor scenes.
The mean of the absolute relative error (AbsRelErr)~\cite{AbsRelErr} between predicted and ground-truth depth is used for evaluation.

\paragraph{Relative camera pose estimation}
Relative camera pose estimation (\Tpose) is about predicting the relative 6D camera poses between the first and last frames of a video sequence. The pose matrix is defined as $P = [R, t]$, where $R$ and $t$ denote the rotation matrix and translation vector, respectively.
The attention readout for action recognition is also utilized for this task. Since the predicted rotation matrix may not be a true rotation matrix in SO(3), the Procrustes algorithm~\cite{bregier2021deep} is applied to the predicted matrix to find the closest true rotation matrix. The readout is trained by minimizing the L2-loss between predicted and ground-truth pose matrices.

We use the RealEstate10k dataset~\cite{zhou2018stereo} which is comprised of indoor and outdoor property videos. The pose annotations are derived from a traditional SfM pipeline, so we rescale camera poses to metric units in order to address scale ambiguities \cite{watson2024controlling}.
The estimated poses are evaluated using mean end-point-error, a metric that measures the mean distance between ground-truth ($P_i$) and estimated ($\hat{P}_i$) pose matrices. More formally, $e_{\text{EPE}}(\hat{P}_i, P_i) = \frac{1}{M}\sum_{j=1}^M \| P_i(Y_j) - \hat{P}_i(Y_j) \|$, where, $\{Y_j\}_{j=1,\dots,M}$ is a set of 3D points selected for metric calculation. In this study, 8 auxiliary points, forming a virtual cube in front of the camera of the first frame, are used for computing $e_{\text{EPE}}$.

\paragraph{Visual correspondence tasks}
Visual correspondence is at the heart of video understanding, as it requires modeling how physical surfaces move and deform over time. In this paper, two correspondence tasks, namely point tracking and box tracking, are selected for evaluation and referred hereafter as \Tpt and \Twaymo, respectively.

The same readout head proposed in MooG \cite{moog} is adopted. Given a set of initial $N$ points (or boxes) at time $t=1$, $q_1{\in}\,\mathbb{R}^{N{\times}D_q}$, and a sequence of observed frames $\{X_t\}_{t=1}^T$, the goal is to predict all future targets, $\{q_t\}$ for $t = 2, \ldots, T$. A latent representation is assigned to $q_1$ by first encoding it with positional encoding followed by an MLP. Then, the latents for $t = 2, \ldots, T$ are generated in a recurrent fashion. At step $t$, they are first predicted by an MLP-based predictor using only the corrected latents at time $t-1$, and then, they are corrected by a transformer, in which the frames are encoded and cross-attended to using the latent predictions as queries to produce the corrections. To generate the final target values, an MLP head is applied to the latents $y_t$. 

The final targets for \Tpt are normalized image coordinates, visibility, and prediction certainty.  A point is considered visible during evaluation only if the model predicts it is visible and has over 50\% confidence in its location. Following MooG \cite{moog}, we use a combined loss function, which includes a Huber loss for location accuracy and Sigmoid Binary Cross Entropy losses for visibility and certainty.  For points that are no longer in the scene, only the visibility loss is applied.
For training the box tracking readout, an L2 loss between the prediction and the normalized box coordinates is used.

We train the point tracking readout head on Kubric {MOVi-E}~\cite{movi-e} labeled with point annotations computed in a similar manner as in~\cite{moog}.
For evaluation, the Perception Test dataset~\cite{perception-test} is used. Sixty four points per frame are sampled and the location of each point in the first frame is used as the query. The average Jaccard (AJ) as in~\cite{moog}, which evaluates both occlusion and position accuracy, is used as performance metric for \Tpt. 
The Waymo Open dataset~\cite{waymo} is used for both training and evaluation of the box tracking readouts and the average IoU (excluding the first frame in the sequence for which ground truth is provided) is used as performance metric.

\section{Experiments}
\label{sec:exp}

We begin our experiments in \cref{sec:exp:qualitative} with qualitative investigations into the features learned by \iwalt and \vwalt.
In \cref{sec:exp:imagevsvideo}, we start our range of quantitative evaluations, comparing representations learned via image and video diffusion pre-training objectives with the help of tasks ranging from pure semantics to spatio-temporal visual understanding.
\Cref{sec:exp:featureextraction} delves further into the design choices involved when extracting features from generative diffusion models.
We conclude our experiments with an analysis on the effect of training budget on feature and generation quality in \cref{sec:exp:fvd} and a comparison with common visual representation models in \cref{sec:exp:baselines}.

\subsection{Qualitative Observations}
\label{sec:exp:qualitative}

\begin{figure}[t]
\centering
\makebox[\linewidth]{
    \newcommand\mypic[1]{
    \includegraphics[width=0.15\linewidth]{imgs/qualitative/pca_one_channel_magma/#1.png}
    }
    \setlength{\tabcolsep}{1pt}
    \begin{tabular}{rcccccc}
        \rotatebox[origin=c]{90}{\hspace{9mm} \footnotesize \centering Frame}
        \mypic{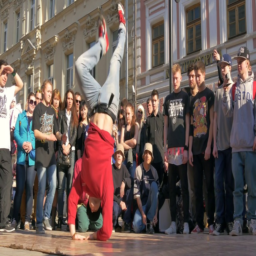} &
        \mypic{} &
        \mypic{} &
        \mypic{} &
        \mypic{} &
        \mypic{} \\[-7mm]
        \rotatebox[origin=c]{90}{\hspace{9mm} \footnotesize \centering \iwalt}
        \mypic{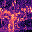} &
        \mypic{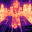} &
        \mypic{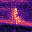} &
        \mypic{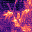} &
        \mypic{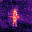} &
        \mypic{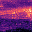} \\[-8mm]
        \rotatebox[origin=c]{90}{\hspace{9mm} \footnotesize \centering \vwalt}
        \mypic{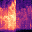} & 
        \mypic{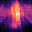} & 
        \mypic{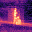} & 
        \mypic{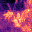} & 
        \mypic{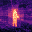} &
        \mypic{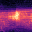} \\[-9mm]
        \rotatebox[origin=c]{90}{\hspace{9mm} \footnotesize \centering Flow}
        \mypic{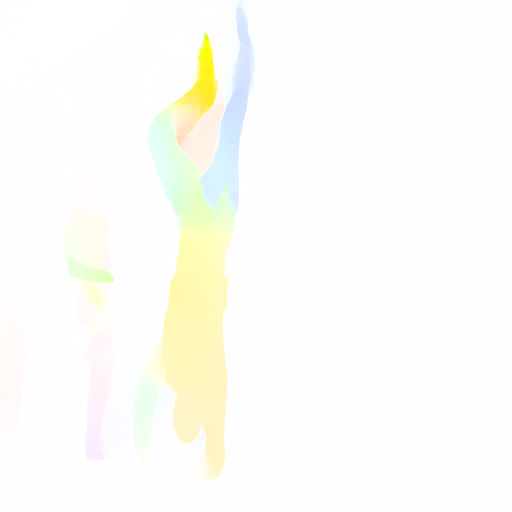} &
        \mypic{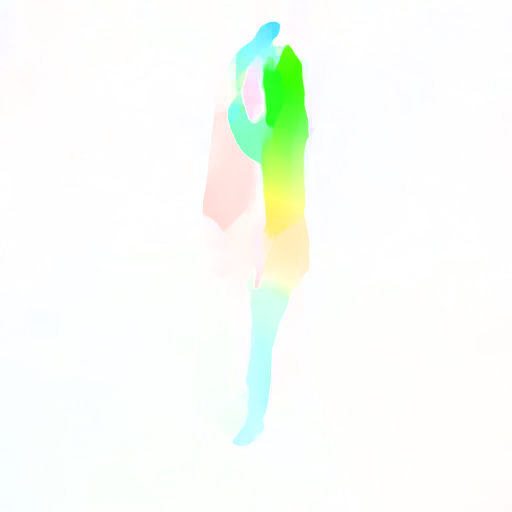} &
        \mypic{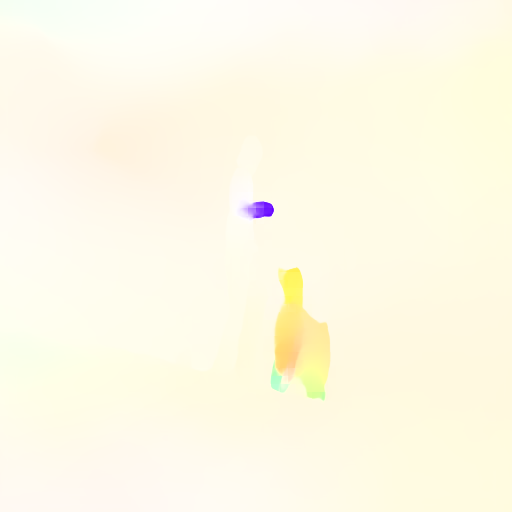} &
        \mypic{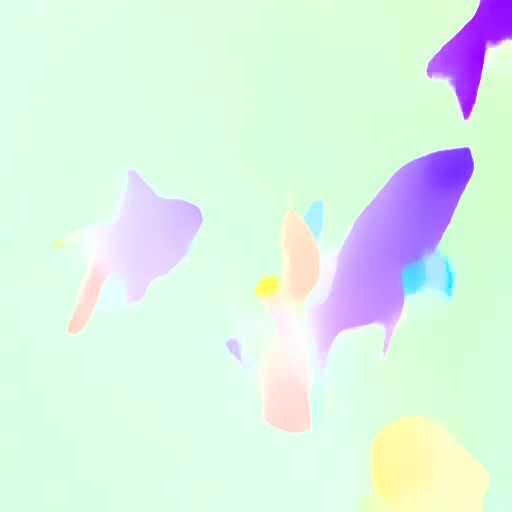} &
        \mypic{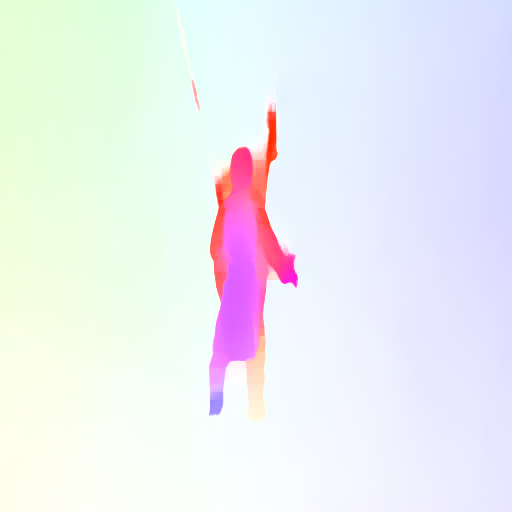} &
        \mypic{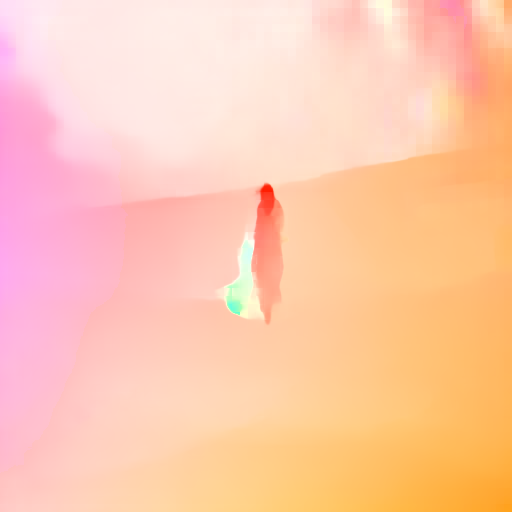} \\[-6mm]
    \end{tabular}
}
\setlength{\belowcaptionskip}{-5pt}
\caption{
\textbf{Feature visualization} --
We show the major PCA component for the two models across a range of DAVIS videos.
While \iwalt is sensitive to semantically important areas of the scene (\eg, \emph{all} people in the second column), \vwalt is much more sensitive to the areas that experience motion within the video (\eg, only the wrestlers in the same video).
}
\label{fig:qualitative1}
\end{figure}

Visualizing latent features in deep neural networks is challenging due to their high dimensionality \cite{Mahendran_2015_CVPR}.
To simplify this analysis, we extract the principal component with the largest eigenvalue from the features at late layers of the \iwalt and \vwalt models.
This allows us to inspect the most salient features in the latent space.

\Cref{fig:qualitative1} visualizes the key feature activations from the first frame of various videos in the DAVIS dataset \cite{davis}, highlighting how \iwalt and \vwalt differ in representing motion. Both models emphasize salient regions, such as people and foreground objects, indicating a shared bias towards key elements in the scene. However, a closer look reveals a key distinction: \iwalt often fails to differentiate between moving and static individuals (\eg, see the second column, where all people are highlighted), while \vwalt selectively focuses on regions exhibiting motion, as confirmed by alignment with optical flow maps.

\begin{figure}[t]
\centering
\newcommand\mypic[1]{
\includegraphics[width=0.18\linewidth]{imgs/qualitative/motion/#1.png}
}
\setlength{\tabcolsep}{0.5pt}
\begin{tabular}{ccccc}
    \mypic{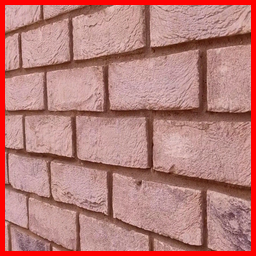} & 
    \mypic{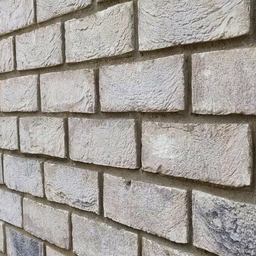} &
    \mypic{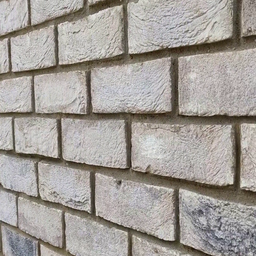} &
    \mypic{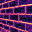} &
    \mypic{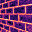} \\[1mm]
    \mypic{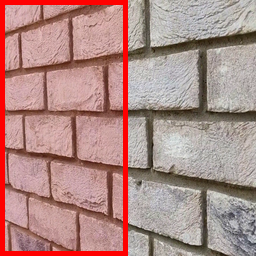} &
    \mypic{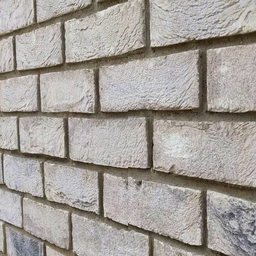} &
    \mypic{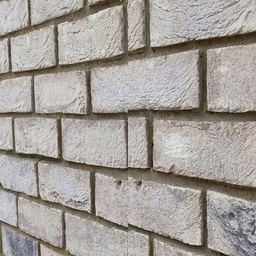} &
    \mypic{brick-iwalt} &
    \mypic{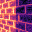} \\[1mm]
    \mypic{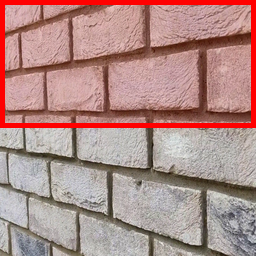} & 
    \mypic{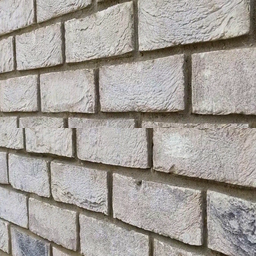} &
    \mypic{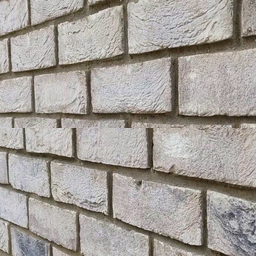} &
    \mypic{brick-iwalt} &
    \mypic{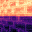} \\[1mm]
    \mypic{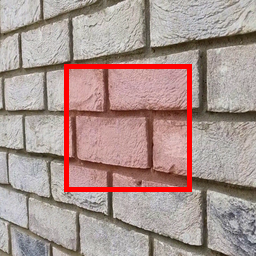} & 
    \mypic{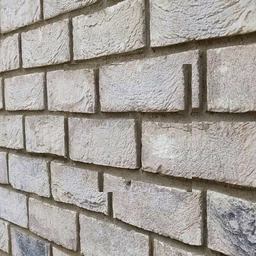} &
    \mypic{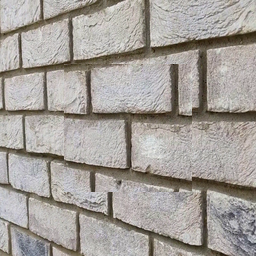} &
    \mypic{brick-iwalt} &
    \mypic{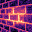} \\[1mm]
    \cmidrule(lr){1-3} \cmidrule(lr){4-4} \cmidrule(lr){5-5} \\[-4mm]
    \multicolumn{3}{c}{Video frames} & \iwalt& \vwalt\\
\end{tabular}
\vspace{-1mm}
\caption{
\textbf{Feature visualization for different motions} --
In the 4 brick videos, only the marked portion (highlighted in red) is played, while the rest remains frozen. We visualize tokens from the first, identical frame. As an image model, \iwalt consistently produces the same feature, while \vwalt shows high sensitivity to moving areas, reflected in the major principal component.
}
\vspace{-4mm}
\label{fig:qualitative2}
\end{figure}

To investigate this observation, a video of a brick wall was manipulated by freezing a region of pixels while displacing a subset within it as shown in \cref{fig:qualitative2}.
Unlike \iwalt, which consistently produces the same features across frames due to its image-based nature, \vwalt shows motion sensitivity, indicated by strong activations in the principal component map. This comparison highlights the ability of \vwalt to distinguish dynamic areas within static scenes, suggesting its suitability for applications that prioritize motion detection and precise differentiation of moving objects.

\subsection{Video versus image diffusion}
\label{sec:exp:imagevsvideo}

Our initial visual inspections of the learned model features pave the way for an objective assessment.
We begin with a key comparison that lies at the heart of our quantitative evaluation.
\Cref{fig:teaser} presents a comparison between the performance of \iwalt, the model trained for image generation, and the performance of the same model architecture trained for video generation (\vwalt) across a range of readout tasks described in \cref{sec:method:probing_framework}.
For a meaningful comparison across tasks, we show the relative performance change $(x_V-x_I)/x_I$, where $x_I$ and $x_V$ denote the absolute performance of the image and video models, respectively.

\emph{\vwalt consistently outperforms \iwalt across all tasks, though the figure reveals a striking range in the extent of this superiority.}
Perhaps unsurprisingly, improvements are small on the purely semantic image classification tasks \Tplaces (+0.6\%) and \Timagenet (+1.8\%), while we see a substantial increase in performance on \Twaymo (+23\%), \Tpose (+60\%) and \Tpt (+68\%) -- tasks that benefit greatly from a deeper understanding of space and motion.

Remarkably, the video training objective improves performance on the image classification task \Tinat (+11\%).
On the action classification tasks, it is interesting to see that the delta is considerable on \Tssv (+42\%), but much more subtle on Kinetics (+8\% on \Tkfour, +12\% on \Tkseven).
This follows the general consensus that the Kinetics tasks primarily measure appearance understanding, while \Tssv is significantly more sensitive to motion understanding \cite{vjepa, sevilla2021only}.

The substantial improvement on \Tpt (+68\%) could be unexpected, given that image diffusion models have been shown to excel at point correspondence \cite{tang2023, hedlin2024unsupervised}.
We attribute this to the fact that point \emph{tracking} is primarily sensitive to accurate localization of \emph{spatial} points in the scene, while point \emph{correspondence} primarily measures semantic understanding.
The videos presented by \citet{hedlin2024unsupervised} illustrate this challenge, demonstrating that the model struggles to distinguish between different instances of the same class (\eg, ears of the same person).

\begin{figure*}[t]
\centering
\includegraphics[width=\linewidth]{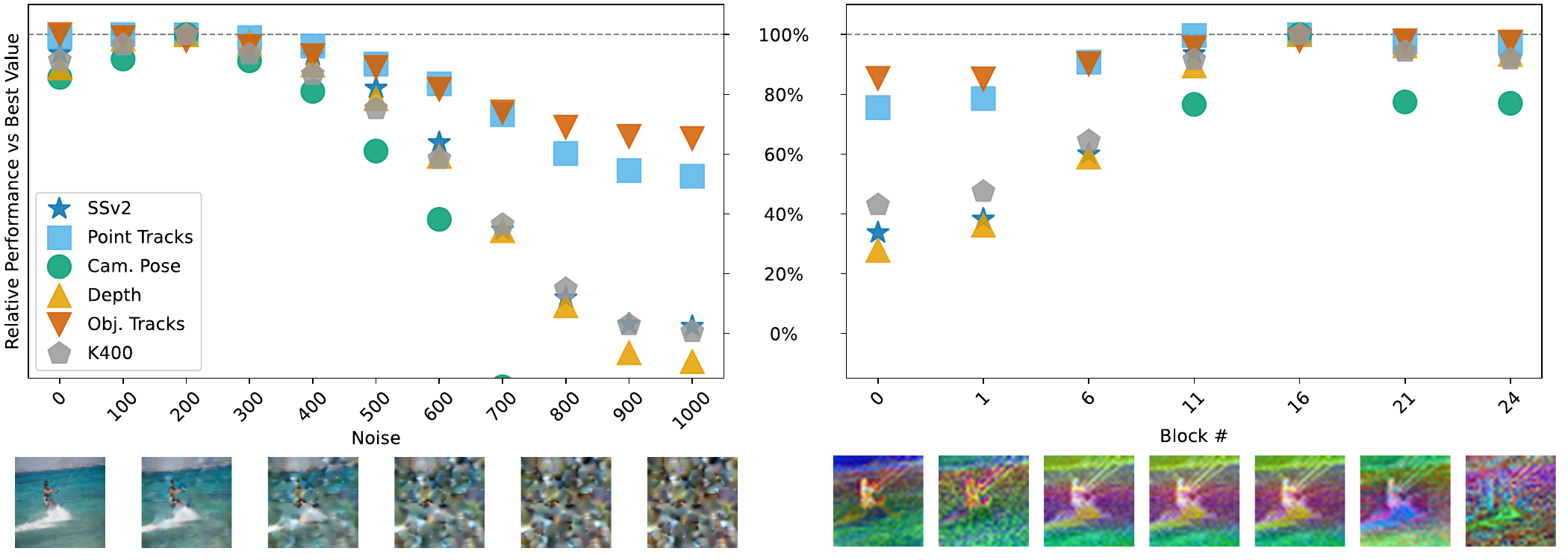}
\vspace{-4mm}
\caption{
    \textbf{Influence of Noise and Block Choice on Readout Performance} --
    Relative change in downstream task performance when probing different noise levels (left, fixed block $l=16$) and intermediate WALT blocks (right, fixed noise $t=200$). Values below -10\% are excluded for clarity.  Optimal performance is generally observed with noise between 0 and 200 and blocks 11-16.  Example noisy images (left) and PCA visualizations (right) are shown below the plots.
}
\label{fig:noise_and_blocks}
\end{figure*}

\subsection{Feature extraction from generative diffusion}
\label{sec:exp:featureextraction}

Trained for denoising, generative diffusion models offer flexibility in feature extraction due to the lack of a dedicated ``feature extraction'' mode.
A crucial consideration is determining both the appropriate time step and the best network block for feature extraction.
To this end, we evaluate the features extracted using various noise levels and transformer blocks within the model across all tasks in \cref{fig:noise_and_blocks}.

\noindent \textbf{Noise level.}
As shown in \cref{fig:noise_and_blocks} (left), introducing high levels of noise generally diminishes downstream task performance. However, a small amount of noise (200) leads to the best results for most tasks, with the exception of \Tpt (best at 100) and \Twaymo (best at 0). This aligns with the intuition that lower-level tasks like tracking are more susceptible to noise, while higher-level tasks benefit from subtle amounts of it.
We find that the image model \iwalt behaves similarly (see the Appendix).
Related works in the literature have investigated this question, though exclusively on image diffusion models, and the consensus is that small amounts of noise have been found helpful for downstream performance, \eg classification \cite{mukhopadhyay2023, xiang2023} or correspondence tasks \cite{luo2023dhf, tang2023}.

\noindent \textbf{Model block.}
Next, we investigate where in the model to extract features from.
The WALT model consists of a tokenizer followed by $L=24$ transformer blocks, see \cref{sec:method:walt}.
As seen in \cref{fig:noise_and_blocks} (right), we find that the best representations are located at a depth of approximately 2\,/\,3 \ within the model for all tasks. A notable deviation from this is the \Tpt task, which performs slightly better at an earlier block.
This finding suggests an implicit separation of the model into encoder and decoder components, leading to the best representations being learned at their intersection.

\subsection{Pre-training budget and generation quality}
\label{sec:exp:fvd}
\vspace{1mm}

Prior work has investigated the relationship between reconstruction and downstream performance \cite{balestriero2024learning} and found that the most informative features are learned towards the end of the training schedule.
To explore this relationship for diffusion models, we train a \vwalt model and capture checkpoints at regular intervals throughout the training process.
Each checkpoint is then evaluated on two fronts:
the effectiveness of its learned features on the downstream tasks, and training progress, as measured by the training loss and Fréchet Video Distance (FVD) \cite{unterthiner2018towards}.

\begin{figure}[t]

\includegraphics[width=\linewidth]{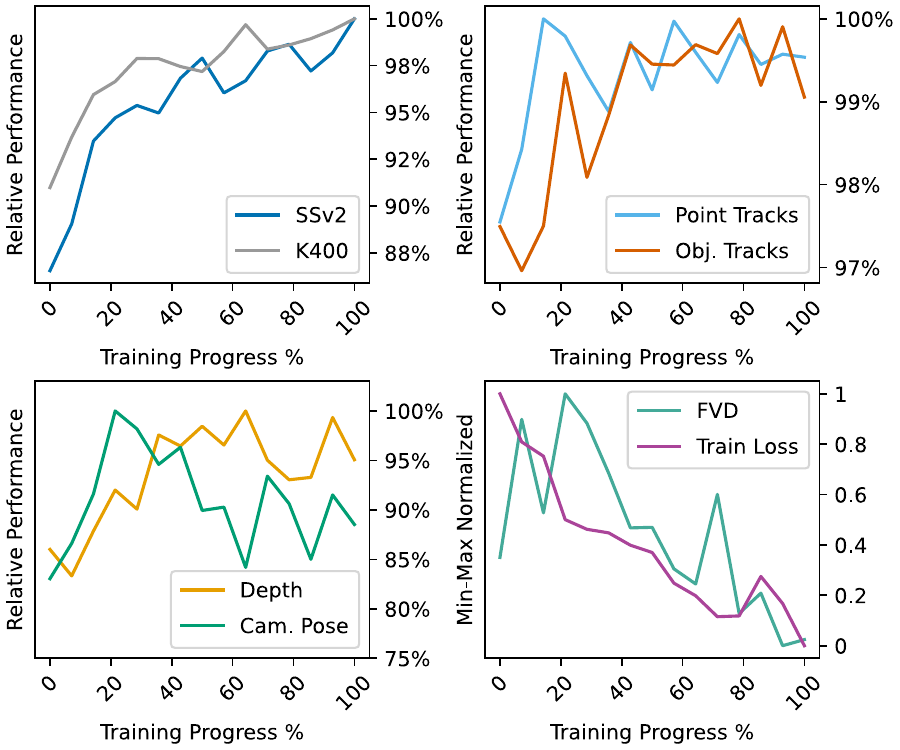}
\vspace{-2mm}
\setlength{\belowcaptionskip}{-10pt}
\caption{
\textbf{Impact of pre-training progress on downstream task performance} -- Recognition tasks generally improve with longer training, while tasks like tracking and depth estimation show optimal performance at earlier stages.  Performance is evaluated across a range of tasks and compared to training loss and Fréchet Video Distance (FVD).
}
\label{fig:training_progress}
\end{figure}

\Cref{fig:training_progress} illustrates the per-task performance of models trained using checkpoints from various stages of the pre-training process.  Interestingly, even early checkpoints (representing 20\% of total pre-training progress) demonstrate a relative performance exceeding 90\% on several tasks. As expected, performance on recognition tasks exhibits a consistent upward trend with continued training. Conversely,  tracking and depth estimation tasks appear to achieve optimal performance at earlier stages. Notably, camera pose estimation performance shows a decline after the 26\% training mark, suggesting potential overfitting or a shift in learned representations that negatively impacts this specific task.

\subsection{Comparisons with visual representation models}
\label{sec:exp:baselines}

To conclude our experiments, we investigate the scaling behavior of the \vwalt model, and compare its performance at sizes 284\,M, as in all other experiments, and 1{.}9\,B, with standard visual representation learning models in the same frozen readout setting. 
We choose representative models from different self-supervised methods:
contrastive DINOv2 \cite{caron2021emerging},
image-text alignment SigLIP \cite{chen2023pali},
pixel reconstruction MAE\cite{mae},
and feature reconstruction model JEPA \cite{ijepa}.
We also include their video extensions VideoMAE \cite{tong2022videomae} and V-JEPA \cite{vjepa} to further explore the differences between models trained on images \vs videos.

\begin{figure}[t]
\centering
\vspace{-2mm}
\includegraphics[
width=\linewidth]{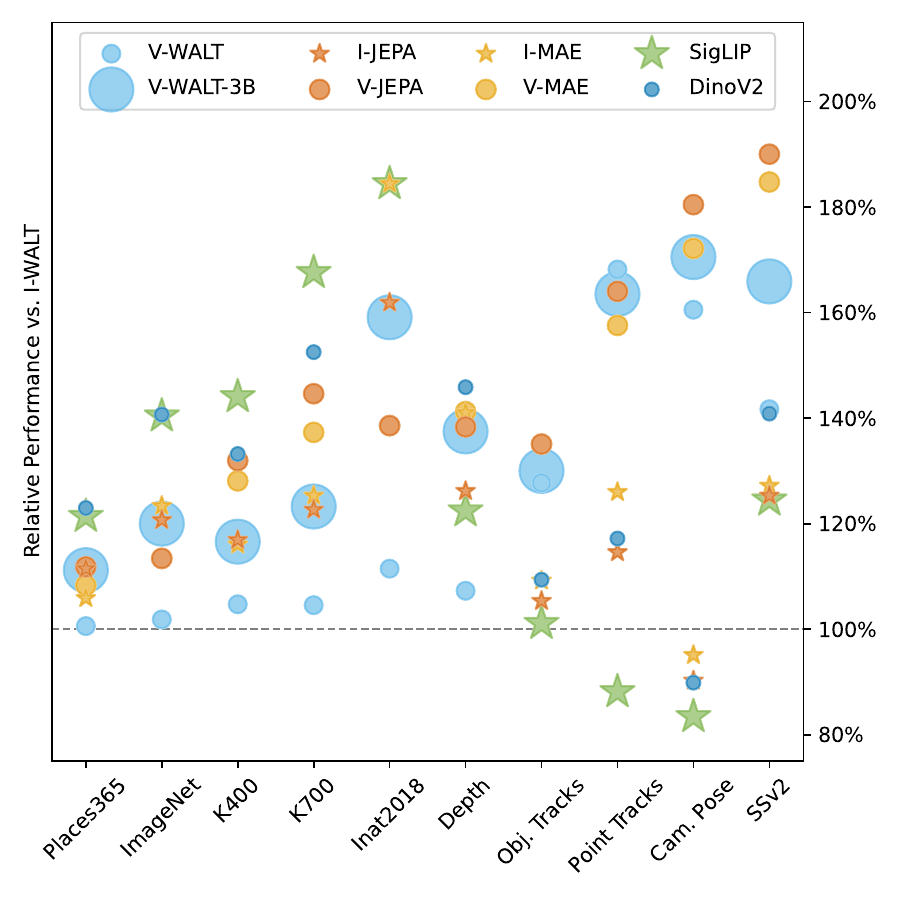}
\vspace{-5mm}
\caption{
\textbf{Relative performance of different SSL models compared to \iwalt.}  We evaluated the variants of different self-supervised models: \vwalt, JEPA, MAE, SigLIP and Dinov2 across 10 image and video tasks using attention probing.\\
\textcolor{gray}{- - -}: \iwalt baseline. $\bigstar$:\;Image models. $\bigcirc$:\;Video models.\\ The size of the makers indicates model size.
}
\vspace{-2.5mm}
\label{fig:baselines}
\end{figure}

In \cref{fig:baselines}, we use the performance of \iwalt as baseline (100\%) and plot the relative performance of other models.
Scaling \vwalt to 1{.}9\,B significantly improves the performance on most tasks, except on \Tpt where the smaller model does marginally better.
The most obvious boosts are on image and video classification, and classification tasks with a large number of classes (\Tkseven and \Tinat) tend to benefit more from the increased model size compared to those with fewer classes (\Tkfour and \Timagenet).
While \vwalt is competitive with the other video models on depth and motion understanding, it is dominated by SigLIP and DinoV2 on the more semantic tasks, revealing a core weakness of generative diffusion models in our readout setting.

\section{Discussion}
\label{sec:conclusion}

In this work, we systematically compared the same model architecture trained for video versus image generation, analyzing the performance of their latent representations on various downstream tasks.
Results show that video diffusion models consistently outperform their image counterparts, especially for tasks that require motion or spatial understanding.
We further analyzed features extracted from different layers and with varying noise levels, as well as the effect of model size and training budget on representation and generation quality.
This work marks the first direct comparison of video and image diffusion objectives for visual understanding, offering insights into the role of temporal information in representation learning.

Our study has several avenues for future work.
Firstly, we limited our study to a single model architecture (WALT \cite{walt}) for a clean comparison that would not be possible for other models that differ considerably between image and video architectures (\eg, SD and SVD), see \cref{sec:method:walt}.
Potential extensions of this work could explore these comparisons across a wider range of unified model architectures as they become available, providing a more comprehensive understanding of the representational power of video and image diffusion models.

Secondly, our investigation primarily focused on the performance of these models on visual understanding tasks.
Future research could delve deeper into the intersection of generative capabilities and representation learning, potentially exploring how the quality of generated images and videos influences and is influenced by the learned representations.
This could lead to new insights and techniques for improving both the generative and representational capabilities of these models.

We believe this study contributes to the ongoing exploration of video and image diffusion models, and we hope our findings inspire further research into their potential for visual understanding and beyond.

\balance

\section*{Acknowledgements}
\label{sec:acknowledgements}
We thank Agrim Gupta and José Lezama for helpful discussions on the WALT model and the 4DS team for providing access to the evaluation setup \cite{s4dpaper}.

{
    \small
    \bibliographystyle{ieeenat_fullname}
    \bibliography{main}
}

\clearpage
\appendix
\setcounter{page}{1}
\maketitlesupplementary

This supplementary material provides further analysis of \iwalt, including ablation studies in \cref{app:sec:iwalt_ablations}, using the same setup as \vwalt. Tables illustrating the performance metrics of the WALT model for image and video tasks are reported in \cref{app:sec:baseline_results}, which correspond to the same values used to calculate the relative performance metrics in \cref{fig:baselines} of the main paper. We also present visualizations of depth estimation and box tracking predictions for both models in \cref{app:sec:qualitative_results}. Details of the datasets and readouts are explained in \cref{app:sec:datasets}, while training settings are described in \cref{app:sec:training_details}.

\section{\iwalt ablations}
\label{app:sec:iwalt_ablations}

To evaluate the image-pretrained model, \iwalt (introduced in \cref{sec:method:walt_variants}), we replicate the ablation study from \cref{sec:exp:featureextraction} for the video-pretrained counterpart, \vwalt.

\begin{figure*}[h]
\centering
\includegraphics[width=\linewidth]{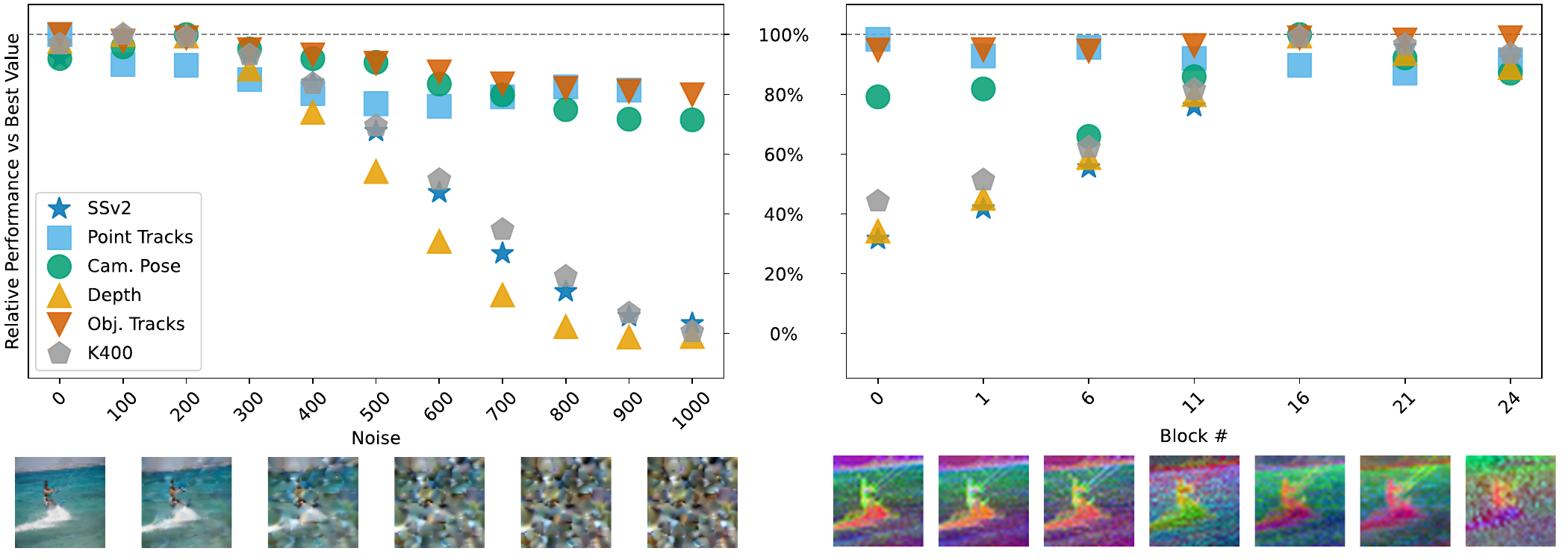}
\vspace{-8mm}
\caption{
    \textbf{Influence of Noise and Block Choice on Readout Performance of \iwalt} --
    Relative change in downstream task performance when probing different noise levels (left, fixed block $l=16$) and intermediate WALT blocks (right, fixed noise $t=200$). Values below -10\% are excluded for clarity.  Optimal performance is generally observed with noise timesteps between 0 and 200 and blocks 11-16.  Example noisy images (left) and PCA visualizations (right) are shown below the plots.
}
\label{fig:iwalt_noise_and_blocks}
\end{figure*}

\paragraph{Noise level}
\Cref{fig:iwalt_noise_and_blocks} (left) illustrates the readout performance of \iwalt at different noise levels.  Following the noise level ablation for \vwalt, the block for feature extraction is set to $l=16$. Similar to what was reported for \vwalt in \cref{sec:exp:featureextraction},  \iwalt benefits from addition of moderate noise levels (between 0 and 200 timesteps).

\paragraph{Model block}
We analyzed the downstream task performance of different layers as shown in \cref{fig:iwalt_noise_and_blocks} (right). As in the model block ablation for \vwalt, the noise timestep is set to $t=200$. Most layers exhibit performance comparable to \vwalt, but point tracking shows a notable difference.  For this task, the earlier layers of the model achieve the highest accuracy. This suggests that the features relevant for point tracking are learned early in the I-WALT transformer.

\begin{figure}[b] 
\includegraphics[width=\linewidth]{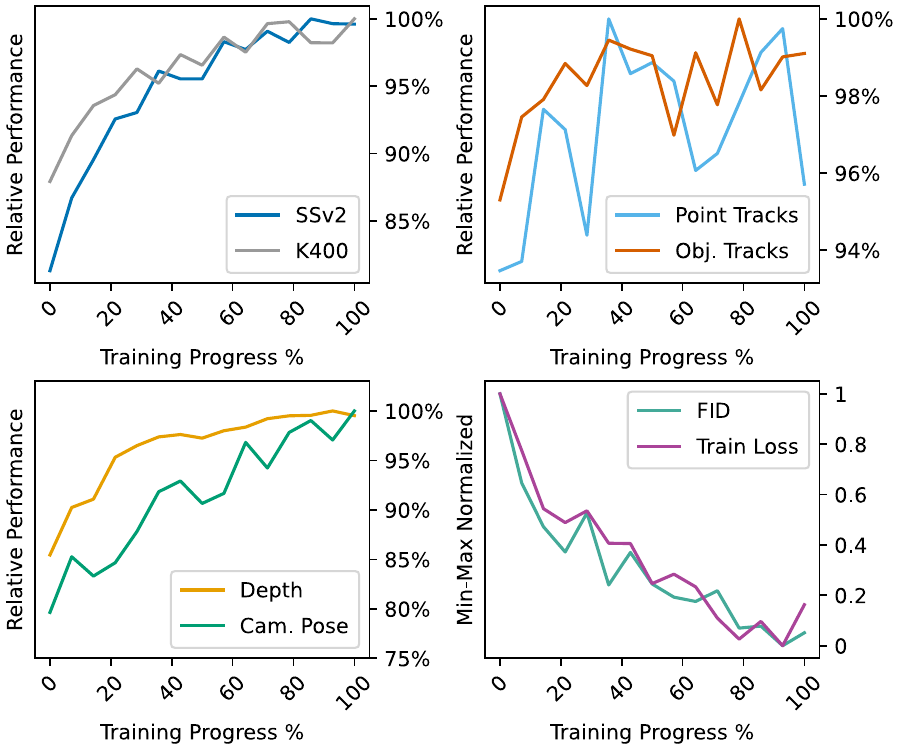}
\vspace{-6mm}
\caption{
\textbf{Impact of pre-training progress on downstream task performance of \iwalt} -- Recognition tasks generally improve with longer training, while tasks like tracking and depth estimation show optimal performance at earlier stages.  Performance is evaluated across a range of tasks and compared to training loss and Fréchet Inception Distance (FID).%
}
\label{fig:iwalt_training_progress}
\end{figure}

\paragraph{Training budget}
\Cref{fig:iwalt_training_progress} shows the training progress of \iwalt using a setup similar to the one described in \cref{sec:exp:fvd}.  We observe a comparable behavior to \vwalt, except for camera pose estimation, where the performance of \iwalt improves with longer training. This suggests that \iwalt does not overfit in this setting, which is further supported by the moderate to strong positive correlations between performance, loss, and FID values over training time, as shown in \cref{tab:iwalt_fid_loss_metric_corr}. Additionally, we include the values for \vwalt in \cref{tab:vwalt_fvd_loss_metric_corr}, which demonstrate a similar correlation pattern, except for camera pose estimation, as mentioned in \cref{sec:exp:fvd}.

\begin{table}[hb]
\centering

\setlength{\tabcolsep}{2pt}
\begin{tabular}{lccccccc}
\toprule
& \Tssv
& \Tkfour
& \Tkseven
& \emph{PointT.} %
& \emph{Cam.\,P.} %
& \Tscannet
& \emph{Obj.\,T.} %
\\
\midrule
\multicolumn{7}{l}{\textit{Pearson coefficient}} \\
Loss
& 0.96
& 0.93
& 0.96
& 0.64
& 0.90
& 0.94
& 0.66\\
FID
& 0.96
& 0.91
& 0.95
& 0.69
& 0.90
& 0.94
& 0.72\\
\midrule
\multicolumn{7}{l}{\textit{Spearman rank correlation coefficient}} \\
Loss
& 0.92
& 0.87
& 0.94
& 0.51
& 0.89
& 0.96
& 0.44\\
FID
& 0.93
& 0.86
& 0.94
& 0.50
& 0.91
& 0.96
& 0.47\\
\bottomrule
\end{tabular}
\caption{
\textbf{Generation quality \vs performance for \iwalt} --
The first row presents Pearson correlation coefficients between downstream performance metrics and training loss. The second row displays the same correlation metric, but calculated between downstream performance and generation quality, as measured by FID. The third and fourth row correspond to similar results with the Spearman correlation coefficient.
}
\label{tab:iwalt_fid_loss_metric_corr}
\end{table}

\begin{table}[hb!]
\centering

\setlength{\tabcolsep}{2pt}
\begin{tabular}{lccccccc}
\toprule
& \Tssv
& \Tkfour
& \Tkseven
& \emph{PointT.} %
& \emph{Cam.\,P.} %
& \Tscannet
& \emph{Obj.\,T.} %
\\
\midrule
\multicolumn{8}{l}{\textit{Pearson coefficient}} \\
Loss
& 0.94
& 0.93
& 0.95
& 0.62
& 0.11
& 0.76
& 0.84\\
FVD
& 0.45
& 0.42
& 0.41
& 0.20
& -0.55
& 0.46
& 0.47\\
\midrule
\multicolumn{8}{l}{\textit{Spearman rank correlation coefficient}} \\
Loss
& 0.94
& 0.89
& 0.97
& 0.27
& -0.07
& 0.61
& 0.72\\
FVD
& 0.63
& 0.70
& 0.66
& 0.27
& -0.50
& 0.45
& 0.52\\
\bottomrule
\end{tabular}
\caption{
\textbf{Generation quality \vs performance for \vwalt} -- The first row presents Pearson correlation coefficients between downstream performance metrics and training loss. The second row displays the same correlation metric, but calculated between downstream performance and generation quality, as measured by FVD~\cite{unterthiner2018towards}. The third and fourth row correspond to similar results with the Spearman correlation coefficient.}

\label{tab:vwalt_fvd_loss_metric_corr}
\end{table}

\section{Baselines results}
\label{app:sec:baseline_results}
The performance metrics of \iwalt and \vwalt for downstream tasks are provided in \cref{tab:image_baselines} and \cref{tab:video_baselines}, respectively. These tables correspond to the values used to generate the relative performance metrics presented in \cref{fig:baselines} of the main paper. For example, accuracy is used for \Tssv action recognition. See~\cref{sec:method:probing_framework} for the full list of metrics used for each downstream task.  

\section{Qualitative results}
\label{app:sec:qualitative_results}

\begin{figure*}[p]
\centering
\newcommand\mypic[1]{
\includegraphics[height=0.145\linewidth, width=0.145\linewidth ]{imgs/qualitative/depth/#1.png}
}
\setlength{\tabcolsep}{0.5pt}
\begin{tabular}{ccccc}
    \mypic{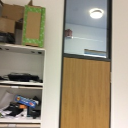} &
    \mypic{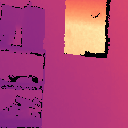} &
    \mypic{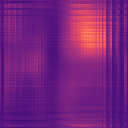} &
    \mypic{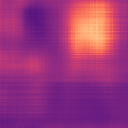} &
    \mypic{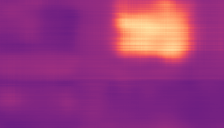} \\
    \mypic{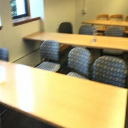} &
    \mypic{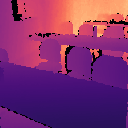} &
    \mypic{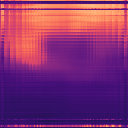} &
    \mypic{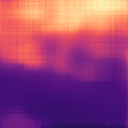} &
    \mypic{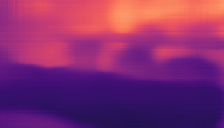} \\
    \mypic{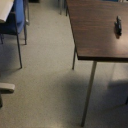} &
    \mypic{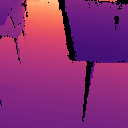} &
    \mypic{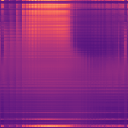} & 
    \mypic{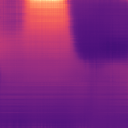} & 
    \mypic{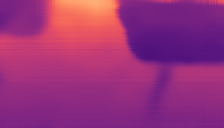} \\
    \mypic{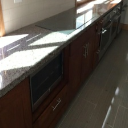} &
    \mypic{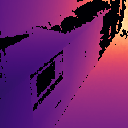} &
    \mypic{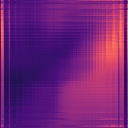} &
    \mypic{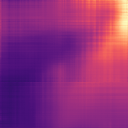} &
    \mypic{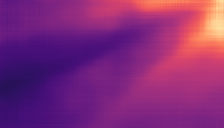} \\
    RGB & GT & \iwalt & \makecell{\vwalt \\ 284M} & \makecell{\vwalt \\ 1.9B} \\
\end{tabular}
\vspace{-1mm}
\caption{
\textbf{Depth predictions from \iwalt and \vwalt} -- RGB, Ground Truth, and Predictions of \iwalt and \vwalt (284M and 1.9B) models. The results of \vwalt 1.9B are resized to a square aspect ratio for visualization purposes.
}
\label{fig:qualitative_depth}
\end{figure*}

\begin{figure*}[p]
\centering
\newcommand\mypic[1]{
\includegraphics[height=0.145\linewidth, width=0.145\linewidth ]{imgs/qualitative/obj_tracks/#1.png}
}
\setlength{\tabcolsep}{0.3pt}
\begin{tabular}{rccccc}
    \rotatebox[origin=c]{90}{\hspace{20mm} \footnotesize \centering GT}
    \mypic{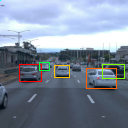} &
    \mypic{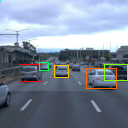} &
    \mypic{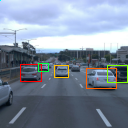} &
    \mypic{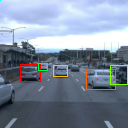} &
    \mypic{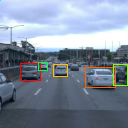} \\[-10mm]
    \rotatebox[origin=c]{90}{\hspace{22mm} \footnotesize \centering \iwalt}
    \mypic{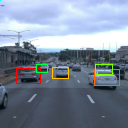} &
    \mypic{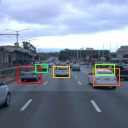} &
    \mypic{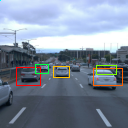} &
    \mypic{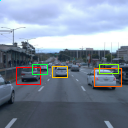} &
    \mypic{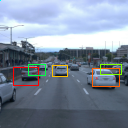} \\[-14mm]
    \rotatebox[origin=c]{90}{\hspace{22mm} \footnotesize \centering \vwalt 284M}
    \mypic{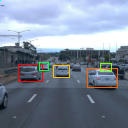} &
    \mypic{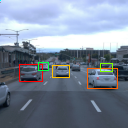} &
    \mypic{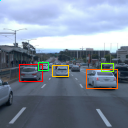} &
    \mypic{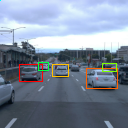} &
    \mypic{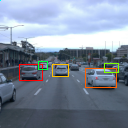} \\[-18mm]
    \rotatebox[origin=c]{90}{\hspace{22mm} \footnotesize \centering \vwalt 1.9B}
    \mypic{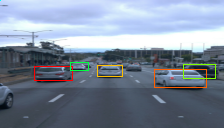} &
    \mypic{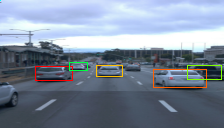} &
    \mypic{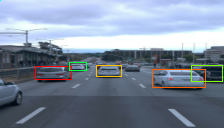} &
    \mypic{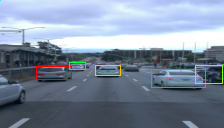} &
    \mypic{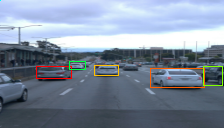} \\[-16.5mm]
    \hline\\[-2mm]
    \rotatebox[origin=c]{90}{\hspace{20mm} \footnotesize \centering GT}
    \mypic{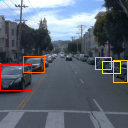} &
    \mypic{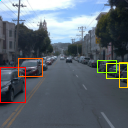} &
    \mypic{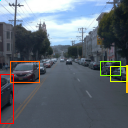} &
    \mypic{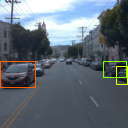} &
    \mypic{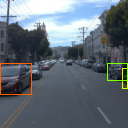} \\[-10mm]
    \rotatebox[origin=c]{90}{\hspace{22mm} \footnotesize \centering \iwalt}
    \mypic{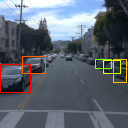} &
    \mypic{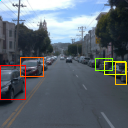} &
    \mypic{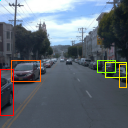} &
    \mypic{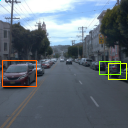} &
    \mypic{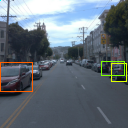} \\[-14mm]
    \rotatebox[origin=c]{90}{\hspace{22mm} \footnotesize \centering \vwalt 284M}
    \mypic{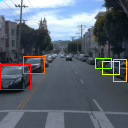} &
    \mypic{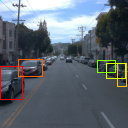} &
    \mypic{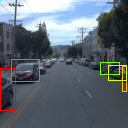} &
    \mypic{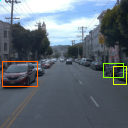} &
    \mypic{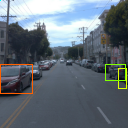} \\[-18mm]
    \rotatebox[origin=c]{90}{\hspace{22mm} \footnotesize \centering \vwalt 1.9B}
    \mypic{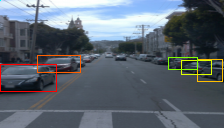} &
    \mypic{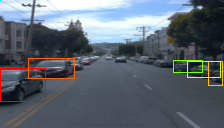} &
    \mypic{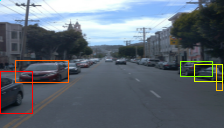} &
    \mypic{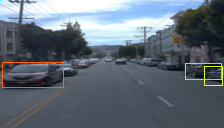} &
    \mypic{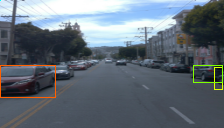} \\[-16mm]
\end{tabular}
\vspace{-1mm}
\caption{
\textbf{Object Tracks predictions from \iwalt and \vwalt} -- Ground Truth, and Predictions of \iwalt and \vwalt (284M and 1.9B) models. The results of \vwalt 1.9B are resized to a square aspect ratio for visualization purposes.
}
\vspace{-4mm}
\label{fig:qualitative_obj_tracks}
\end{figure*}

\begin{figure*}[t]
\centering
\newcommand\mypic[1]{
\includegraphics[height=0.145\linewidth, width=0.145\linewidth ]{imgs/qualitative/ssv2/#1.png}
}
\setlength{\tabcolsep}{0.5pt}
\begin{tabular}{cccccc}
    Video &
    \mypic{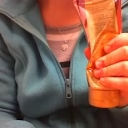} &
    \mypic{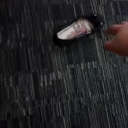} &
    \mypic{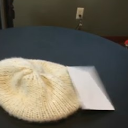} &
    \mypic{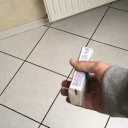} &
    \mypic{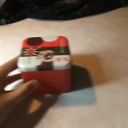}\\[2mm]
    GT &
    \makecell{Squeezing\\something} &
    \makecell{Pretending\\to pick\\something up} &
    \makecell{Dropping\\something next\\to something} &
    \makecell{Pretending to\\throw something} &
    \makecell{Pushing something\\from left\\to right} \\[8mm]
    \iwalt &
    \makecell{Holding\\something} &
    \makecell{Picking\\something up} &
    \makecell{Turning the camera\\upwards while\\filming something} &
    \makecell{Pretending to\\throw something} &
    \makecell{Pushing something\\so that it\\slightly moves} \\[8mm]
    \vwalt 284M &
    \makecell{Squeezing\\something} &
    \makecell{Pretending\\to pick\\something up} &
    \makecell{Dropping\\something next\\to something} &
    \makecell{Pretending to\\throw something} &
    \makecell{Pushing something\\from left\\to right} \\[8mm]
    \vwalt 1.9B &
    \makecell{Squeezing\\something} &
    \makecell{Pretending\\to pick\\something up} &
    \makecell{Dropping\\something next\\to something} &
    \makecell{Pretending to\\throw something} &
    \makecell{Pushing something\\from left\\to right} \\[10mm]
\end{tabular}
\vspace{-5mm}
\caption{
\textbf{Action recognition predictions from \Tssv using \iwalt and \vwalt} -- The top row displays a single frame from a dataset sample. Frames were manually selected to best showcase the corresponding label.
}
\vspace{-4mm}
\label{fig:qualitative_ssv2}
\end{figure*}

In order to qualitatively assess the performance of \iwalt compared to \vwalt, we include prediction outputs of all WALT models for the tasks of monocular depth prediction (\cref{fig:qualitative_depth}), box tracking (\cref{fig:qualitative_obj_tracks}) and action recognition (\cref{fig:qualitative_ssv2}).
As demonstrated in \cref{sec:exp:imagevsvideo}, \vwalt consistently outperforms \iwalt across all tasks, particularly those requiring spatiotemporal understanding. This is evident in the enhanced accuracy of \vwalt predictions, such as depth maps exhibiting greater pixel correspondence, even with the \vwalt 218M model, which has the same number of parameters as \iwalt.  Similarly, \vwalt shows superior object tracking performance, while \iwalt struggles to accurately estimate the size and location of bounding boxes.  The temporally rich input data enables \vwalt to achieve higher accuracy in action recognition. For instance, \vwalt correctly classifies the action of ``squeezing something'' (first column), which occurs midway through the video clip, whereas \iwalt fails to do so.

\section{Datasets and readout details}
\label{app:sec:datasets}

 A summary of the readout architectures and their corresponding parameter counts for the image and video tasks is provided in \cref{table:readout_modules_image} and \cref{table:readout_modules_video}, respectively.

\begin{table*}[t]
\centering
\begin{tabular}{lccc}
\toprule
Model & \Tplaces $\uparrow$ & \Timagenet $\uparrow$ &  \Tinat $\uparrow$ \\
\midrule
\multicolumn{4}{l}{\textit{Methods pretrained on image tasks}} \\
I-JEPA-600M           & 0.514 & 0.732 & 0.421 \\
ImageMAE-600M         & 0.488 & 0.749 & 0.479 \\
SigLIP 1.7B           & 0.559 & 0.852 & 0.479 \\
DinoV2-300M           & 0.567 & 0.854 & 0.726 \\
I-WALT 284M           & 0.461 & 0.607 & 0.260 \\
\midrule
\multicolumn{4}{l}{\textit{Methods pretrained on video tasks}} \\
V-JEPA-300M     & 0.509 & 0.678 & 0.349 \\
V-JEPA-600M     & 0.515 & 0.688 & 0.360 \\
VideoMAEv1-600M & 0.499 & 0.643 & 0.321 \\
V-WALT 284M     & 0.464 & 0.618 & 0.290 \\
V-WALT 1.9B     & 0.512 & 0.728 & 0.414 \\
\bottomrule
\end{tabular}
\caption{
\textbf{Comparison with state-of-the-art methods on image recognition tasks} -- All results presented here were obtained using the same training and evaluation protocol with frozen backbones and trainable readouts.
}
\label{tab:image_baselines}
\end{table*}

\begin{table*}[t]
\centering
\begin{tabular}{lccccccc}
\toprule
Model & \Tkfour $\uparrow$ & \Tkseven $\uparrow$ & \Tscannet $\downarrow$ & \makecell{\Twaymo $\uparrow$} & SSv2 $\uparrow$ & \makecell{\Tpose $\downarrow$} & \makecell{\Tpt $\uparrow$} \\
\midrule
\multicolumn{8}{l}{\textit{Methods pretrained on image tasks}} \\
I-JEPA-600M                 & 0.617 & 0.485 & 0.147 & 0.483 & 0.451 & 2.299 & 0.515 \\
ImageMAE-600M               & 0.612 & 0.496 & 0.117 & 0.501 & 0.458 & 2.197 & 0.566 \\
SigLIP 1.7B  & 0.760 & 0.663 & 0.154 & 0.464 & 0.448 & 2.442 & 0.396 \\
DinoV2-300M                 & 0.702 & 0.604 & 0.108 & 0.502 & 0.507 & 2.307 & 0.526 \\
I-WALT 284M                 & 0.527 & 0.396 & 0.199 & 0.459 & 0.360 & 2.095 & 0.449 \\
\midrule
\multicolumn{8}{l}{\textit{Methods pretrained on video tasks}} \\
V-JEPA-300M     & 0.685 & 0.557 & 0.132 & 0.629 & 0.658 & 0.507 & 0.733 \\
V-JEPA-600M     & 0.696 & 0.572 & 0.123 & 0.620 & 0.684 & 0.409 & 0.737 \\
VideoMAEv1-600M & 0.675 & 0.543 & 0.117 & 0.620 & 0.665 & 0.583 & 0.708 \\
V-WALT 284M     & 0.552 & 0.414 & 0.185 & 0.586 & 0.510 & 0.826 & 0.756 \\
V-WALT 724M     & 0.571 & 0.445 & 0.151 & 0.587 & 0.547 & 0.814 & 0.741 \\
V-WALT 1.9B     & 0.615 & 0.488 & 0.124 & 0.597 & 0.597 & 0.617 & 0.735 \\
\bottomrule
\end{tabular}
\caption{
\textbf{Comparison with state-of-the-art methods on video recognition, depth estimation, tracking, and camera pose estimation tasks} -- All results presented here were obtained using the same training and evaluation protocol with frozen backbones and trainable readouts.
}
\label{tab:video_baselines}
\end{table*}

\begin{table*}[h]
\centering
\begin{tabular}{c|c c}
\hline
\textbf{Task} & \textbf{Architecture} & \textbf{Number of parameters} \\ \hline
\Timagenet & 
\begin{tabular}[c]{@{}l@{}}
\texttt{CrossAttention(} \\
\texttt{\quad qkv\_size=768,} \\
\texttt{\quad num\_heads=12)} \\
\\ 
\texttt{Dense(output\_size=1000)}
\end{tabular} & 7,678,184 \\ \hline
\Tplaces & 
\begin{tabular}[c]{@{}l@{}}
\texttt{CrossAttention(} \\
\texttt{\quad qkv\_size=768,} \\
\texttt{\quad num\_heads=12)} \\
\\ 
\texttt{Dense(output\_size=365)}
\end{tabular} & 7,189,869 \\ \hline
\Tinat & 
\begin{tabular}[c]{@{}l@{}}
\texttt{CrossAttention(} \\
\texttt{\quad qkv\_size=768,} \\
\texttt{\quad num\_heads=12)} \\
\\ 
\texttt{Dense(output\_size=8142)}
\end{tabular} & 13,170,382 \\ \hline
\end{tabular}
\caption{Architecture details and number of parameters for the image classification readouts.}
\label{table:readout_modules_image}
\end{table*}

\begin{table*}[htb]
\centering
\begin{tabular}{c|c c}
\hline
\textbf{Task} & \textbf{Architecture} & \textbf{Number of parameters} \\ \hline
\Tssv Action Recognition & 
\begin{tabular}[c]{@{}l@{}}
\texttt{CrossAttention(} \\
\texttt{\quad qkv\_size=768,} \\
\texttt{\quad num\_heads=12)} \\
\\ 
\texttt{Dense(output\_size=174)}
\end{tabular} & 7,042,990 \\ \hline
\Tkfour Action Recognition & 
\begin{tabular}[c]{@{}l@{}}
\texttt{CrossAttention(} \\
\texttt{\quad qkv\_size=768,} \\
\texttt{\quad num\_heads=12)} \\
\\ 
\texttt{Dense(output\_size=400)}
\end{tabular} & 11,975,056 \\ \hline
\Tkseven Action Recognition & 
\begin{tabular}[c]{@{}l@{}}
\texttt{CrossAttention(} \\
\texttt{\quad qkv\_size=768,} \\
\texttt{\quad num\_heads=12)} \\
\\ 
\texttt{Dense(output\_size=700)}
\end{tabular} & 12,282,556 \\ \hline
Relative camera pose estimation & 
\begin{tabular}[c]{@{}l@{}}
\texttt{CrossAttention(} \\
\texttt{\quad qkv\_size=256,} \\
\texttt{\quad num\_heads=8)} \\
\\
\texttt{Dense(output\_size=12)}
\end{tabular}  & 1,650,444 \\ \hline
Monocular depth prediction & 
\begin{tabular}[c]{@{}l@{}}
\texttt{CrossAttentionTransformer(} \\
\texttt{\quad qkv\_size=512,} \\
\texttt{\quad num\_heads=2,} \\
\texttt{\quad mlp\_size=512,} \\
\texttt{\quad num\_layers=1)}
\end{tabular} & 3,284,353 \\ \hline
Waymo Object Tracking & 
\begin{tabular}[c]{@{}l@{}}
\texttt{CrossAttentionTransformer(} \\
\texttt{\quad qkv\_size=512,} \\
\texttt{\quad num\_heads=8,} \\
\texttt{\quad mlp\_size=2048,} \\ 
\texttt{\quad num\_layers=3)} \\
\\ 
\texttt{predictor=MLP(} \\
\texttt{\quad hidden\_size=512,} \\
\texttt{\quad output\_size=512)} \\
\\
\texttt{output=MLP(} \\
\texttt{\quad hidden\_size=512,} \\
\texttt{\quad output\_size=6)}
\end{tabular} & 12,931,462 \\ \hline
Point Tracking & 
\begin{tabular}[c]{@{}l@{}}
\texttt{CrossAttentionTransformer(} \\
\texttt{\quad qkv\_size=512,} \\
\texttt{\quad num\_heads=8,} \\
\texttt{\quad mlp\_size=2048,} \\
\texttt{\quad num\_layers=3)} \\
\\
\texttt{predictor=MLP(} \\
\texttt{\quad hidden\_size=512,} \\
\texttt{\quad output\_size=512)} \\
\\ 
\texttt{output=MLP(} \\
\texttt{\quad hidden\_size=512,} \\
\texttt{\quad output\_size=4)}
\end{tabular} & 12,897,668 \\ \hline
\end{tabular}
\caption{\textbf{Readout heads setup} -- Architecture details and number of parameters for the readouts of the video tasks.}
\label{table:readout_modules_video}
\end{table*}

\subsection{Image classification}

\paragraph{Dataset}
Image classification results are reported on \Timagenet~\cite{imagenet}, \Tplaces~\cite{places365}, and \Tinat~\cite{inaturalist}.

\Timagenet ~\cite{imagenet} is a large-scale dataset containing 1,000 object and animal categories, with 1,281,167 images for training and 50,000 images for validation. For training on \Timagenet, we follow the traditional augmentation scheme consisting of inception cropping and random horizontal flipping. For evaluation, a $224\times224$ center crop is extracted from each image whose shorter edge is first resized to 256.

\Tplaces is a scene classification dataset that contains images with buildings, landscapes, and other everyday scenarios. It includes 1.8 million training images and 36,500 validation images across 365 scene classes. 

\Tinat contains 437{,}513 images for training and 24{,}426 images for validation, featuring 8,142 classes of visually similar species across a wide range of taxonomic groups, such as plants, animals, fungi, and insects. The dataset also exhibits heterogeneous image quality due to diverse camera sources and exhibits a substantial class imbalance.

The same data augmentation and preprocessing steps used for \Timagenet are applied to both \Tinat and \Tplaces. For all the image classification datasets, the original data splits are used for training and evaluating the classification readout.

\paragraph{Readout}
In the case of \vwalt, an image is replicated across the 17 temporal input channels and the model features are averaged before passing them to the readout. \iwalt does not require any image replication and the model features are directly passed to the readout.

Adopting the approach of V-JEPA~\cite{vjepa}, we utilize a cross-attention block with a learnable query token to extract class information from the model features. This token attends to the features within the cross-attention block, and its output is fed into a linear classifier for class prediction. The readout is trained with the softmax cross-entropy  loss.

\subsection{Action recognition}

\paragraph{Dataset}
Something-Something-V2 (\Tssv)~\cite{ssv2} is a large-scale dataset consisting of short videos (2-6 seconds at 12 frames per second) depicting diverse human actions with everyday objects.  The dataset is specifically designed for fine-grained understanding of human hand gestures, focusing on subtle actions, such as placing objects into containers. Something-Something-V2 encompasses 174 categories with 168,913 samples in the training set and 24,777 in the validation set.

Kinetics-400 (\Tkfour) is a large-scale dataset of YouTube videos designed for human action recognition, encompassing object manipulation, human-object interaction, and body movements. Kinetics-400 consists of 246,245 training videos and $\sim$20K validation videos with an average duration of 10 seconds. All video clips are labeled into 400 classes.

Kinetics-700 (\Tkseven) is an extension of Kinetics-400. The data collection pipeline between the two datasets differs in how action classes are sourced, how videos are matched with classes, and the human verification process. Kinetics-700 contains 545,317 training videos and 35,000 validation videos across 700 fine-grained action classes.

For both training and evaluation of the readout, 17 frames are sampled from each video to generate the model input. A stride of 2 is used for \Tssv, while a stride of 1 is used for \Tkfour and \Tkseven. For  all  the  action recognition datasets, the original data splits are used for training and evaluating the readout.

\paragraph{Readout}
The same attention readout used for image classification and described above is also used for action recognition.

\subsection{Monocular depth prediction}
\paragraph{Dataset}
For this work, we utilize the train and validation splits of the ScanNet dataset~\cite{scannet}, comprising 1,201 and 312 videos respectively. The dataset offers high-resolution RGB frames (1296x968) in diverse indoor environments and corresponding depth frames (640x480) captured with an RGB-D system. The input to the WALT model is obtained by sampling 17 consecutive frames from the ScanNet videos. During training, the starting frame is chosen randomly, while for evaluation, sampling always begins at frame 0. %

\paragraph{Readout}
We use the readout from~\cite{SRT}, which applies cross-attention with spatial coordinates as queries to each frame independently. The readout is trained using an L2 loss between predicted and ground truth depth maps.

\subsection{Relative camera pose estimation}
\paragraph{Dataset}
RealEstate10K~\cite{zhou2018stereo} is a dataset of property walkthrough videos with intrinsic and extrinsic camera parameters using Structure from Motion (SfM). The clips were gathered from YouTube and typically feature smooth camera movement with minimal camera roll or pitch. The original splits of the dataset are used, which consist of roughly 10 million training frames from 6,500 videos and 1 million test frames from 696 videos.

\paragraph{Readout}
The input of the readout is formed by concatenating the video representations of the first and last frame of the video sequences.  These are then processed via cross-attention with learned latent vectors and a linear layer to produce 12-dimensional vectors representing SE(3) pose transformations, which correspond to a $3\times3$ rotation matrix and a $3\times1$ translation vector. The predicted rotation matrix is refined using the Procrustes algorithm~\cite{bregier2021deep} to ensure it represents a valid SO(3) rotation before metric evaluation. Training is performed by minimizing the L2 loss between predicted and ground-truth pose matrices.

\subsection{Visual correspondence -- Point tracking}
\paragraph{Dataset}
The Perception Test dataset~\cite{perception-test} was specifically designed to evaluate the perception and reasoning skills of multimodal video models. It was filmed by around 100 participants worldwide and contains perceptually interesting situations. In this paper, the Perception Test dataset is used to evaluate the point tracking task. Specifically, the validation set is employed, which comprises 73 real-world videos, averaging 722 frames in length, with multiple point tracks annotated using the same protocol as in~\cite{moog}. Each point is visible in approximately 480 frames. The first 17 frames of each video are sampled with a stride of 4 to generate the model input.

The point tracking readout head is trained with the training set of the Kubric MOVi-E dataset~\cite{movi-e}, which contains 97,500 synthetic 24-frame videos, each depicting scenes with 10-20 static and 1-3 dynamic objects rendered against photorealistic backgrounds. The camera in these videos moves on a straight line at a constant velocity, always pointed towards the origin of the scene. 

\paragraph{Readout}
To build the readout input, pretrained model features are first interpolated in the temporal dimension to match the number of video frames. The interpolated features and a set of query points become the readout input. During training, 17 frames and 64 point tracks are randomly sampled from each video. Then, a random crop with an area between 30\% and 100\% of the original frame and an aspect ratio between 0.5 and 2.0 is extracted. The crops are then resized to $128\times 128$. Query points are selected exclusively from the first frame.

For evaluation, we sample the first 17 frames with a stride of 4 from each video and use 64 point tracks. As in~\citep{moog}, our evaluation takes the first visible point track as the query, and discards frames preceding its appearance.

The readout head employs an iteratively applied cross-attention transformer, maintaining a 512-dimensional latent state for each point track between frames. As in~\citep{moog}, this state is initialized from query point positions using a Fourier positional encoding followed by a two-layer MLP. The transformer comprises three layers of cross-attention with eight heads and a key/value size of 512. At each step, it uses the latent state as queries to attend to the frame features generated by the pretrained model. A two-layer MLP predicts the position, visibility, and uncertainty of each point track in each frame using the corresponding latent states as input. The loss function is a combination of a Huber loss for location accuracy and Sigmoid Binary Cross Entropy losses for visibility and certainty.  Points that have exited the scene contribute only to the visibility loss.

\subsection{Visual correspondence -- Box tracking}
\paragraph{Dataset}
We leverage the Waymo Open Dataset~\cite{waymo}, utilizing the high-resolution (1280x1920) RGB video data captured at 10 fps. This dataset, recorded from Waymo vehicles in urban and suburban settings, includes 2D and 3D bounding box annotations. We use the 2D bounding boxes for loss calculation and metric evaluation. The training and validation sets comprise 798 and 202 samples, respectively, each 20 seconds in duration. The same data splits are used for training and evaluating the box tracking readout.

\paragraph{Readout}
Consistent with prior work~\cite{moog}, for both training and evaluation, we downsample videos to $256\times384$ resolution at 5 fps, and then extract a central 256x256 spatial crop and a random 17-frame temporal crop. Bounding boxes smaller than 0.5\% of the first sampled frame area are discarded, and a maximum of 25 boxes are retained per sample.

The same attention readout used for point tracking and described above is also used for box tracking, only differing in the predictions. The position $x_{min}$, $x_{max}$, $y_{min}$, $y_{max}$ of query boxes is predicted for box tracking. The box tracking readout is trained using an L2 loss between the predicted and normalized box coordinates. As in point tracking, the pretrained model features are interpolated in the temporal dimension to match the number of video frames. 

\section{Training settings}
\label{app:sec:training_details}
\Cref{tab:experiments_setup} summarizes the hyperparameters used to train the readout heads for each task described in \cref{app:sec:datasets}. We use the AdamW optimizer with a cosine learning rate decay schedule, initialized with a linear warmup over 1,000 steps (from 0 to 3e-4), and subsequently decaying to 1e-7. Batch sizes are 32 for video tasks, 512 for \Tplaces and \Tinat, and 64 for \Timagenet.
\begin{table*}[htb]
\centering
\begin{tabular}{lcccc}
\toprule
 & Places365 &  iNat2018 & ImageNet & \makecell{Video \\ Datasets} \\
\cline{2-5}\\
batch size & 512 & 512 & 64 & 32 \\
\makecell[l]{training steps} & 10,000 & 10,000 & 200,182 & 40,000 \\
optimizer & adamw & adamw & adamw & adamw \\
\hspace{1mm} $\epsilon$ & 1e-8 & 1e-8 & 1e-8 & 1e-8 \\
\hspace{1mm} $\beta_1$ & $\num{0.9}$ & $\num{0.9}$ & $\num{0.9}$ & $\num{0.9}$ \\
\hspace{1mm} $\beta_2$ & $\num{0.999}$ & $\num{0.999}$ & $\num{0.999}$ & $\num{0.999}$ \\
\hspace{1mm} weight decay & 1e-4 & 1e-4 & 1e-4 & 1e-4 \\
lr schedule & cosine & cosine & cosine & cosine \\ 
\hspace{1mm} init. lr & 0 & 0 & 0 & 0 \\
\hspace{1mm} peak lr & 3e-4 & 3e-4 & 3e-4 & 3e-4 \\
\hspace{1mm} warmup steps & 1,000 & 1,000 & 1,000 & 1,000 \\
\hspace{1mm} lr end value & 1e-7 & 1e-7 & 1e-7 & 1e-7 \\
\bottomrule
\end{tabular}
\caption{
\textbf{Readout hyperparameters} -- Parameters used to train the image and video readout heads using a frozen pretrained WALT model.
}
\label{tab:experiments_setup}
\end{table*}

\end{document}